\documentclass{article}

\usepackage{microtype}
\usepackage{graphicx}
\usepackage{amsmath} 
\usepackage{amssymb} 
\usepackage{cancel} 
\usepackage{subfigure}
\usepackage{booktabs} 

\usepackage{hyperref}

\newcommand\norm[1]{\left\lVert#1\right\rVert} 

\usepackage[accepted]{icml2020_arxiv}

\icmltitlerunning{Learning Continuous Treatment Policy and Bipartite Embeddings for Matching with Heterogeneous Causal Effects}

\begin{document}

\twocolumn[
\icmltitle{Learning Continuous Treatment Policy and Bipartite Embeddings for Matching with Heterogeneous Causal Effects}




\begin{icmlauthorlist}
\icmlauthor{Will Y. Zou}{ubmi}
\icmlauthor{Smitha Shyam}{ubmi}
\icmlauthor{Michael Mui}{ubmi}
\icmlauthor{Mingshi Wang}{ubmi}
\icmlauthor{Jan Pedersen}{ubai}
\icmlauthor{Zoubin Ghahramani}{ubai}
\end{icmlauthorlist}

\icmlaffiliation{ubmi}{Uber Michelangelo, San Francisco, USA}
\icmlaffiliation{ubai}{Uber AI, San Francisco, USA}

\icmlcorrespondingauthor{Will  Zou}{will.zou@uber.com} 

\icmlkeywords{Deep Learning, Causal Inference, Machine Learning, Heterogeneous Treatment Effect, Treatment Effect Estimation}

\vskip 0.3in
]



\printAffiliationsAndNotice{} 
\begin{abstract} 
Causal inference methods are widely applied in the fields of medicine, policy, and economics. Central to these applications is the estimation of treatment effects to make decisions. Current methods make binary yes-or-no decisions based on the treatment effect of a single outcome dimension. These methods are unable to capture continuous space treatment policies with a measure for \emph{intensity}. They also lack the capacity to consider the \emph{complexity} of treatment such as matching candidate treatments with the subject. 


We propose to formulate the effectiveness of treatment as a parametrizable model, expanding to a multitude of treatment intensities and complexities through the continuous policy treatment function, and the likelihood of matching. Our proposal to decompose treatment effect functions into effectiveness factors presents a framework to model a rich space of actions using causal inference. 

We utilize deep learning to optimize the desired holistic metric space instead of predicting single dimensional treatment counterfactual. This approach employs a population-wide effectiveness measure and significantly improves the overall effectiveness of the model. 

We demonstrate the superior performance of our algorithms. When using generic continuous space treatments and matching architecture, we observe a 41\% improvement upon prior art with cost effectiveness and 68\% improvement upon a similar method in average treatment effect. The algorithms captures subtle variations in treatment space, structures the efficient optimizations techniques, and opens up the arena for many applications.

\end{abstract} 

\section{Introduction} 
\label{sec:intro} 

Given a large set of patients and medications, how does one recommend the best medication and its dosage to each patient through observation of treatment effects on a multitude of outcomes? 

Past work in causal inference provides solutions through estimation of treatment effect~\cite{kunzel2017meta}~\cite{shalit2017estimating}~\cite{nie2017quasi}. The effect on a treatment subject is addressed with a \emph{`counterfactual'}~\cite{pearl2009causality} argument of \emph{what if the subject was given a different treatment} quantified by the Individual Treatment Effect (ITE) function~\cite{kunzel2017meta}~\cite{nie2017quasi}.

This approach can estimate counterfactual outcomes to measure effectiveness of treatments. However, the approach requires a prohibitively large number of models, since a subject can be matched with exponentially large number of actions or choices,  each leading to a multitude of outcomes. The complexity introduces the Curse of Dimensionality~\cite{friedman1997bias, bishop2006pattern}, resulting in data sparsity in the large parameter space. The solution calls for a generalization to continuous space treatment variables, and accounting for shared latent factors.  

We propose a solution to capture both the complexity of actions and outcomes. At the heart of this solution is Bayesian decomposition combined with deep learning methods. 

We address the intractability of continuous-space integration of the Average Treatment Effect (ATE) function with Bayesian methods. Instead of integrating over the continuous space, we formulate probability density weighting with Bayesian decomposition. With multiple ATE functions, we construct a combined objective to account for the multitude of outcomes in a joint optimization problem. 

We are the first to propose continuous space treatment policy for causal inference. Instead of evaluating the counterfactual of a binary decision, we assume the treatment is given at continuous levels such as the dosage of medication or the price of a product. We estimate likelihood of the treatment intensity using continuous distributions and re-weight the prior. This method allows us to answer questions such as `what is the optimal level of treatment for a subject?' and `how effective is the treatment if given at 53\% as opposed to another?'. We leverage deep learning to formulate a model for matching subject to treatment types. Activations from hidden layers in neural networks form embeddings for subject and treatment candidates. The distance measure across embeddings is used as matching score, as applied by learning-to-rank~\cite{huang2013learning}. 

We demonstrate from first principles how the Average Treatment Effect (ATE) functions in causal inference can be decomposed into multiple factors, namely, a continuous policy likelihood, a matching likelihood and a subject instance prior. We also parameterize the model using deep learning techniques and construct the objective using multiple ATE functions. The powerful combination of Bayesian decomposition and deep learning provides a highly flexible and effective framework to optimize for the holistic treatments effectiveness with arbitrary objective functions. 

Our key contributions can be summarized as follows: 




\textbf{Continuous Treatment Policy} - We are the first to incorporate continuous treatment intensity into the causal machine learning framework. We formulate treatment policies as continuous variables and propose probability distributions to model them. The effectiveness measures can then be defined with arbitrary distributions that factor into the objective. The eventual aggregated objective offers training signals to determine the optimal continuous treatment level. 

\textbf{Bipartite Embeddings for Relevance Matching} - Compared with previous research, our proposal makes decisions on subject instance being matched with the types of treatment, rather than only considering subject instance scoring for treatment. We consider the likelihood of match in the perspective of \emph{relevance matching} in bipartite instances, and formulate embedding spaces for both subject instance and matched instance. For example, this can be patients with medications, and customers with products. 

\textbf{Bayesian Decomposition for Chained Causal Factors} - Instead of integrating over a series of latent variables, Bayesian decomposition offers a generic way to incorporate a chain of factors into the aggregated treatment effect function. The holistic objective can eventually be optimized with deep learning. This offers a framework for incorporating additional factors beyond continuous treatment and bipartite matching, making it adaptable to a wide range of application domains. 

\textbf{Deep Learning with Causal Effectiveness Objective} - We derive a causal objective function to optimize for treatment effectiveness. Instead of estimating treatment effects of a single outcome, the objective aggregates contributions from multiple outcome dimensions.  Effectiveness can thus be evaluated with regards to the combined objective. Further, the loss function can be utilized in deep learning models with flexible architectures. We demonstrate superior algorithmic flexibility of this modeling framework evidenced by experimentation. 

\vspace{-0.2cm}
\section{Prior Work} 
\label{sec:prior_work} 

The space of causal inference has been studied in the context of treatment effect estimation ~\cite{shalit2017estimating, kunzel2017meta, nie2017quasi, johansson2016learning, swaminathan2015self, swaminathan2015counterfactual} which uncovers a range of algorithms from \emph{meta-learners}~\cite{kunzel2017meta}, to balancing the representation space~\cite{shalit2017estimating}.  ~\cite{louizos2017causal} discovers hidden confounding factors using techniques such as variational auto-encoders, ~\cite{parbhoo2018causal} investigates from information bottleneck specific to causal algorithms. ~\cite{lim2018forecasting} adopted recurrent networks to study the effects of sequential treatments. There are many applications to precision medicine~\cite{shalit2017estimating}~\cite{lim2018forecasting}. It is worth noting the classic work ~\cite{rubin1974estimating} which proposed a paradigm for treatment effect estimation whereby the outcome of the subject treatment is observed and subsequently used to fit a model for counterfactual estimation. Statistical viewpoints have been taken by ~\cite{kunzel2017meta} to decompose the learning algorithm into composite models with \emph{meta-learners}. Notably, quasi-oracle estimation by~\cite{nie2017quasi} is effective when estimating the treatment effect in a single outcome. Recently, decision trees and random forests~\cite{chen2016xgboost} have been applied~\cite{rzepakowski2012decision} as another mainstream methodology for treatment effect estimation. This includes causal tree,  random forests~\cite{wager2018estimation,athey2016recursive}, boosting~\cite{powers2017some, hill2011bayesian}. 

Most previous methods consider the \emph{Average Treatment Effect (ATE)} of a single treatment, indicated by a binary variable across the population and commonly considered with respect to a single dimension of outcome. Previous approaches are not suited for dealing with continuous treatments, multi-dimensional outcomes, or matching with treatment types. We propose a way to cope with continuous treatment variables and create an architecture for matching treatment types to subjects, integrating into a framework that is generalizable to a chain of factors using Bayesian decomposition. The framework eventually leverages deep learning to optimize for aggregated treatment effect. 

\section{Algorithms} 
\label{sec:algorithms}
\subsection{Problem Statement: Matching Algorithm with Continuous Space Policy} 
Consider when a patient, based on her medical history and symptoms, needs to be matched with a treatment. The treatment is characterized with suitable symptoms and properties. When prescribed, there is a continuous measure for how intense the treatment should be. The goal is match the best treatment, with the optimal treatment intensity. Another example is pricing at an internet company, when a user needs to be matched with any product in the on-line store, and continuous treatment is the price to allocate for the specific user-product match. 

The quasi-oracle estimation algorithm, among other meta-learner algorithms~\cite{kunzel2017meta}, is capable of estimating treatment effects in a multitude of dimensions. However, this is in-efficient due to exponentially large outcome and action spaces. The approach in this paper is to maximize an overall effectiveness measure, as a key theme in applications of causal inference. We propose causal inference paradigm to maximize combined effectiveness of heterogeneous treatments. 



The objective for our algorithm is to identify a \emph{collection of treatment sessions}, each composed of \emph{a pair of subject and treatment candidate}, that achieves the highest treatment effectivness considering many outcomes. To formulate the problem, we introduce a notation of \emph{the overlined treatment effect} $\overline{\tau}^{*}(\mathbf{x}) = E(Y_1 - Y_0 | \mathbf{X} = \{\mathbf{x}^{(i)}\})$, which represents the treatment effect function across the collection of subjects $\{\mathbf{x}^{(i)}\}$. The models will be parametrized by $\mathbf{\theta}$ and effectiveness measures $p_{\mathbf{\theta}}(\mathbf{x}^{(i)})$ per session $i$. The models output the effectiveness measures that correlate with the expectation across the collection, i.e. $~\overline{\tau}^{*}(\mathbf{x}) = \sum_i p(\mathbf{x}^{(i)})Y_1^{(i)} - \sum_j p(\mathbf{x}^{(j)})Y_0^{(j)}$ which can be combined into the same expression using $\overline{\tau}^{*}(\mathbf{x})$. 

Note the critical step to represent the objective in the causal deep learning paradigm as a treatment effect function parametrized by multiple deep learning models. This allows us to represent the eventual objective function as a combination of multiple $\overline{\tau}^{*}(\mathbf{x})$. 

Suppose one tries to solve the following optimization problems: 
\begin{align} 
\label{eq:pstatement_objectives_1} 
\text{maximize} \quad &\overline{\tau}^{*q}(\mathbf{x})(\overline{\tau}^{*r}(\mathbf{x}) - \lambda \overline{\tau}^{*c}(\mathbf{x})) \\ 
\label{eq:pstatement_objectives_2}
\text{minimize} \quad &\frac{\overline{\tau}^{*c}(\mathbf{x})}{\overline{\tau}^{*r}(\mathbf{x})} + \lambda \overline{\tau}^{*m}(\mathbf{x}) 
\end{align} 


The first problem tries to maximize a combined treatment effect of reward minus cost weighted by an overall factor $q$. The second represents a cost efficiency measure, with an extra weighting factor $m$. We aim to optimize for the holistic effectiveness exemplified in these equations. The holistic objective allows us to take all subjects, treatments, and outcomes into consideration. The unconstrained optimization allows us to utilize deep learning to build flexible models and efficiently train with large-scale data. 


\textbf{Bayesian Decomposition with Causal Inference.} Without loss of generality, we include per sample treatment propensity from causal statistics~\cite{lunceford04stratification} and derive $\bar{\tau}$ without superscript as it could extend to both $\bar{\tau}^{*r}$ and $\bar{\tau}^{*c}$. Starting from the fundamental definition of treatment effect: 
\begin{align*} 
\bar{\tau}^{*} &= E(Y_1 - Y_0) = E(Y_1) - E(Y_0) 
\end{align*} 
The treatment effect term can be conditioned on a treatment policy. Given a policy $\mathbb{P}$, we can write: 
\begin{align*} 
\bar{\tau}^{*} &= E(Y_1 - Y_0 | \mathbb{P}) = E(Y_1 | \mathbb{P}) - E(Y_0 | \mathbb{P}) 
\end{align*} 

Different from prior work~\cite{lunceford04stratification}, we differentiate the Policy $\mathbb{P}$ with the treatment cohort indicator $T$, the latter $T$ random variable indicates whether an instance is in the treatment cohort or in the control cohort. Only within the treatment cohort, the Policy $\mathbb{P}$ is applied. The Policy can be optimized to produce the best possible outcome in treatment cohort, when treatment cohort variable is correlated with whether assign instances to control or held-out cohort. The propensity in the context of a possible policy evaluation, is the expected value of treatment cohort indicator. 

From~\cite{lunceford04stratification}, the expected value of outcome of any instance can be written as following equations. This takes Treatment Cohort Indicator and propensities into account: 
\begin{align} 
E(Y_1 | \mathbb{P}) = E(\frac{Y_1T}{e(\mathbf{x})} | \mathbb{P}),  \quad E(Y_0 | \mathbb{P}) = E(\frac{Y_0 (1-T)}{1-e(\mathbf{x})}| \mathbb{P})
\end{align} 
$e(\mathbf{x})$ is a propensity function $E(T=1|\mathbf{X} = \mathbf{x})$. This quantity is estimated given features of the subject instance, thus can be specific per instance with a learnable propensity function fitted with the feature and treatment cohort indicator labels. Detailed proof of the above equation is given for $Y_1$ case: $E(Y_1) = E(\frac{Y_1}{e(\mathbf{x})} e(\mathbf{x})) = E(\frac{Y_1}{e(\mathbf{x})}E(T | \mathbf{X}=\mathbf{x})) = E(\frac{Y_1}{e(\mathbf{x})}E(T | \mathbf{X}=\mathbf{x}, Y_1)) = E(E(\frac{Y_1T}{e(\mathbf{x})} | X, Y_1)) = E(\frac{Y_1T}{e(\mathbf{x})})$\footnote{Third step follows from unconfoundedness assumption~\cite{lunceford04stratification}~\cite{nie2017quasi}.} 

Substitute into the treatment effect definition: 
\vspace{-0.1cm}
\begin{align} 
\label{eq:drm_propensity_1} 
\begin{split} 
\bar{\tau^{*}} &= E(Y_1 - Y_0 | \mathbb{P}) = E(Y_1 | \mathbb{P}) - E(Y_0 | \mathbb{P}) \\ 
&= E(\frac{Y_1T}{e(\mathbf{x})} | \mathbb{P})  - E(\frac{Y_0 (1-T)}{1-e(\mathbf{x})} | \mathbb{P}) \\ 
&= E(E(\frac{Y_1T}{e(\mathbf{x})}|T, \mathbb{P})) - E(E(\frac{Y_0(1-T)}{1-e(\mathbf{x})}|T, \mathbb{P})) \\ 
\end{split} 
\end{align} 
\vspace{-0.1cm}

The last step follows from law of total expectation. We then expand the equation and eliminate zero terms: 
\vspace{-0.1cm}
\begin{align} 
\label{eq:drm_propensity_2} 
\begin{split} 
\bar{\tau}^{*} &= P(T = 1) E(\frac{Y_1T}{e(\mathbf{x})}|T = 1, \mathbb{P}) \\ 
& - P(T = 0) E(\frac{Y_0 (1-T)}{1-e(\mathbf{x})} | T = 0, \mathbb{P}) 
\end{split} 
\end{align} 
\vspace{-0.1cm}


We expand expectation in the following equations. Every match is decomposed into two instances of the items being matched, which denote with random variables $I$ and $J$ indexed by $i$ and $j$. We define $\hat{e}$ to be the overall propensity, or likelihood of being in the treatment cohort across all instances. For each user, the $e(\mathbf{x})$ term is estimated from a pre-trained propensity function. In the last equation, we introduce matches of instances denoted by random variable $M$, and use the tuple notation $M=(I, J)$ and $m = (i, j)$ to indicate each match across two distinct items. 
\vspace{-0.1cm}
\begin{align} 
\label{eq:bayes_weighted_tau} 
\bar{\tau}^{*} &= \hat{e} \sum_{T_{i,j} = 1}\frac{p( I=i, J=j | \mathbb{P}^{(i,j)})}{e(\mathbf{x})} Y_1^{(i,j)} \\ & - (1 - \hat{e}) \sum_{T_{i, j} = 0} \frac{p( I=i, J=j | \mathbb{P}^{(i, j)})}{1-e(\mathbf{x})} Y_0 ^ {(i,j)} \\ 
& = \hat{e} \sum_{T_{m} = 1} \frac{p(M=m) p(\mathbb{P}^{(m)} | M=m )}{e(\mathbf{x})\sum_l p(M=l)p(\mathbb{P}^{(l)} | M=l)} Y_1^{(m)} \\
&  - (1 - \hat{e}) \sum_{T_m = 0} \frac{p(M=m ) p(\mathbb{P}^{(m)} | M = m) }{(1-e(\mathbf{x}))\sum_l p(M=l)p(\mathbb{P}^{(l)} | M = l)}Y_0 ^ {(m)} 
\end{align} 
\vspace{-0.1cm}
\normalsize 
The last two lines are obtained using Bayes Rule to rewrite the posterior $p( I=i, J=j | \mathbb{P}^{(i,j)})$ with $\frac{p(M=m) p(\mathbb{P}^{(m)} | M=m )}{Z} $, prior multiplied by likelihood, divided by the normalization factor or partition function $Z$. 

As stated before, we parameterize the above function, then combine the Treatment Effect Functions $\overline{\tau}^{*}$, we will be able to compose an objective function for the learning algorithm. 

\textbf{Continuous treatment policy model.} The key to defining an objective function composed of conditional Average Treatment Effect (ATE) representations with $\tau^*$ is finding representations for the likelihood $p(\mathbb{P} | M = m)$, the probability of a policy given a match. We solve the problem with a continuous treatment policy problem formulation. This is to say, we assume the policy random variable $\mathbb{P}$ to be a continuous random variable, assuming its its scope to be $\mathbb{P}\in [0, 1]$. $\mathbb{P}$ represents intensity of treatment. The likelihood $p(\mathbb{P} | M = m)$ is defined to be the family of distributions with scope $[0, 1]$ that can be parameterized by $s(\mathbf{x}^{(i)}, \mathbf{y}^{(j)})$: 

\vspace{-0.1cm}
\begin{align} 
p(\mathbb{P}^{(i, j)} | M = m) = \mathbf{\emph{D}}(\mathbb{P}^{(i,j)} | s(\mathbf{x}^{(i)}, \mathbf{y}^{(j)})_{\mathbb{P}}) 
\end{align} 
\vspace{-0.1cm}

For the $s$ quantity, given any specific match, it is then formulated a parameterized regression model: 
\vspace{-0.1cm}
\begin{align} 
\label{eq:bell_mean} 
s(\mathbf{x}^{(i)}, \mathbf{y}^{(j)})_{\mathbb{P}} = f(\mathbf{x}^{(i)}, \mathbf{y}^{(j)}) 
\end{align} 
\vspace{-0.1cm}

\begin{figure}[h] 
  \centering 
  \includegraphics[width=0.75\linewidth]{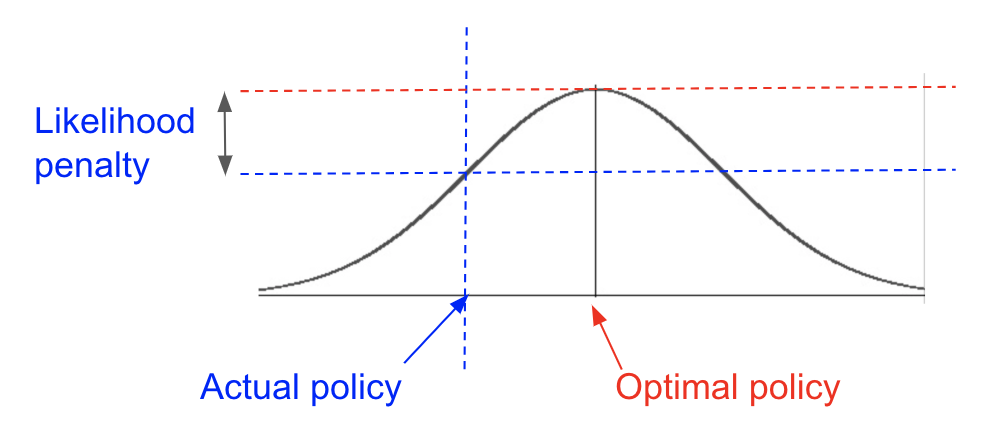} 
  \caption{Illustration of penalty of a sub-optimal policy.} 
  \label{fig:bell_policy_penalty} 
\end{figure}

Feature vectors $\mathbf{x}^{(i)}$ and $\mathbf{y}^{(j)}$ denote features to specify the match subject $i$ and object $j$, e.g. user and product, patient and treatment. The intuition behind this formulation is the regression model determines hyper-parameters of the continuous distribution, which then measures the likelihood of any continuous policy value. The distribution is distinct per match, and offers measure for the goodness of policy values. For example, when $s$ denotes the mean of a bell-shaped distribution, the regression model should gives the optimal treatment intensity $s_{\mathbb{P}}$ for the specific match of subject and object. During training, if the actual data deviates from this optimal value, its likelihood would be penalized with respect to the amount of deviation. This is shown in Figure~\ref{fig:bell_policy_penalty}. Further, functional form of $s$ can be any differentiable regression algorithm, such as a logistic regression, or multi-layer neural network. 

\vspace{-0.1cm}
\vspace{-0.1cm}

For the likelihood function, we can use a distribution whose density function can be specified by the derivative of the sigmoid: 

\vspace{-0.1cm}
\begin{align} 
p(\mathbb{P}^{(i, j)} | M = m) \propto \sigma(\hat{\mathbb{P}}^{(i,j)}) (1 - \sigma (\hat{\mathbb{P}}^{(i,j)})) \\ 
\text{where} \quad \hat{\mathbb{P}}^{(i,j)} = \mathbb{P}^{(i, j)} - s(\mathbf{x}^{(i)}, \mathbf{y}^{(j)})_{\mathbb{P}} 
\end{align} 
\vspace{-0.1cm}

$\sigma$ denotes the sigmoid function and $\sigma (1 - \sigma)$ is its derivative. This formulation gives the likelihood a probabilistic interpretation, such that the cumulative density of the random variable takes forms of the sigmoid\footnote{The distribution has the scope $[0,1]$ and is re-normalized.}, and the probabilistic density takes a bell-shaped form. This means whenever the intensity $\mathbb{P}$ deviates from the optimal, the likelihood function penalizes by lowering its value, positioning the highest value at top of bell curve, determined by Equation~\ref{eq:bell_mean}. 

This likelihood function can also be defined with the \emph{Beta} distribution: 
\vspace{-0.1cm}
\begin{align} 
p(\mathbb{P}^{(i,j)} | M = m) = \mathbf{\emph{B}}(\mathbb{P}^{(i,j)} | s_\alpha(\mathbf{x}^{(i)}, \mathbf{y}^{(j)}), s_\beta(\mathbf{x}^{(i)}, \mathbf{y}^{(j)})) 
\end{align} 
\vspace{-0.1cm}

Where the random variable $\mathbb{P}^{(i,j)}$ assumes a density of a Beta distribution with parameters $\alpha, \beta$ which are parameterized regression function from feature sets $\mathbf{x}$ and $\mathbf{y}$. This second formulation has the interpretation for variable $\mathbb{P}^{(i,j)}$ to be in the range of $[0, 1]$. If both parameters are limited to be above $1.0$, the distribution also takes the bell-shaped form with maximum density at $\frac{\alpha}{\alpha + \beta}$. 

\textbf{Parameterize subject-candidate matching model.} We next parameterize the matching model. 
We can decompose the following equation: 
\vspace{-0.1cm}
\begin{align} 
\label{eq:matching_func} 
p(M = m) =& p (I = i, J = j) \\ =& p(I = i) p(J = j | I = i) \\ \propto& g(\mathbf{x}^{(i)}) h(\mathbf{x}^{(i)}, \mathbf{y}^{(j)}) 
\end{align} 
\vspace{-0.1cm}

A simple form of $g$ is a sigmoid compressed version of arbitrary differentiable regression function, and forms the base instance weighting model. With sigmoid as the non-linearity, the outputs are positive and are well-controlled below $1.0$. 

For the functional form of $h$ we adopt a bipartite embedding formulation. First the input features $\mathbf{x}$ are projected with a neural network to an embedding space for both subject instance, and matching candidate: $\mathbf{e}_i = f_{sp} (\mathbf{x}^{(i)}), \mathbf{e}_j = f_{cp} ( \mathbf{y}^{(j)} )$. In this manner, the embedding space can be learned using eventual effectiveness measures, across the treatment subjects, e.g. patient, as well as the treatment candidates, e.g. medications. The eventual objective helps to learn the embedding sub-space each of the subject and candidate lies in. Also the algorithm for matching here is highly related to learning-to-rank models such as DSSM~\cite{huang2013learning}. Where the candidates are projected first into a vector space before the metric distance is used to define the relevance of terms. Note that we use the same projection function in the definition of $g$. Then, we can formulate the likelihood function: 
\vspace{-0.1cm}

\begin{align} 
\label{eq:matching_func} 
h(\mathbf{x}^{(i)}, \mathbf{y}^{(j)}) &= 1 + cos(\phi) = 1 + \frac{\mathbf{e}_i \cdot \mathbf{e}_j}{\norm{\mathbf{e}_i \cdot \mathbf{e}_j}} \\ 
&= 1 + \frac{f_{sp}(\mathbf{x}^{(i)}))\cdot f_{cp}(\mathbf{y}^{(j)})}{\norm{f_{sp}(\mathbf{x}^{(i)}))\cdot f_{cp}(\mathbf{y}^{(j)})}} 
\end{align} 
\vspace{-0.1cm}
\textbf{Normalization into effectiveness measures.} With the previously defined likelihood functions for both treatment policy and matching, the functions need to be normalized to offer a probabilistic interpretation. 

Concretely, all output measures from distributions and function proposed in the above sections are positive. We sum the scores together to form the partition function then normalize the product of prior and likelihood. This is done for across all possible data instances, or matches across subject and candidate, each of these is indexed by $i,j$. 

Written as $Z(I, J) = \sum_{i,j} p(M=(i,j))p(\mathbb{P}^{(l)} | M = (i,j))$, the partition function normalizes parameterized likelihood. Thus, Equation~\ref{eq:bayes_weighted_tau} can be computed to define the overall objective function with: 
\begin{align}
&p(M = m) = \frac{p(I=i)p(J = j | I = i)}{Z(I,J)} = \frac{g(\mathbf{x}^{(i)})h(\mathbf{x}^{(i)},\mathbf{y}^{(j)})}{\hat{Z}_P(I,J)} \\ 
&p(M = m | \mathbb{P}^{(m)}) = \frac{p(M=m)p_{\mathbf{\emph{D}}}(\mathbb{P}^{(m)} | M=m)}{\hat{Z}_M(M)} 
\end{align} 
\vspace{-0.10cm}

\textbf{Objective Function.} Summing up the parameterized policy model, matching model, and normalization, we can write the eventual average treatment effect function as follows. Note we notate all the parameterized functions with a top bar, and all of them are a differentiable function with respect to parameters. 
\begin{align} 
\label{eq:final_obj} 
\begin{split} 
\bar{\tau}^{*} &= \hat{e} \sum_{T_m =1} \frac{Y^{(m)}_1  \bar{g}(\mathbf{x}^{(i_m)})\bar{h}(\mathbf{x}^{(i_m)}, \mathbf{y}^{(j_m)})\bar{p}_{\mathbf{\emph{D}}}(\mathbb{P}^{(m)} | M=m)}{e(\mathbf{x}^{(i_m)}, \mathbf{y}^{(j_m)})\hat{\bar{Z}}_M(M)\hat{\bar{Z}}_P(I,J)} \\&- (1-\hat{e}) \sum_{T_m = 0} \frac{Y^{(m)}_0 \bar{g}(\mathbf{x}^{(i_m)})\bar{h}(\mathbf{x}^{(i_m)}, \mathbf{y}^{(j_m)})\bar{p}_{\mathbf{\emph{D}}}(\mathbb{P}^{(m)} | M=m)}{(1-e(\mathbf{x}^{(i_m)}, \mathbf{y}^{(j_m)}))\hat{\bar{Z}}_M(M)\hat{\bar{Z}}_P(I,J)} 
\end{split} 
\end{align} 
\vspace{-0.10cm}

Then we substitute the expression of the ATE function back to Equations~\ref{eq:pstatement_objectives_1}~\ref{eq:pstatement_objectives_2} to obtain the eventual objective function. Since all functions are differentiable, we can obtain parameter gradients and optimize the objective. 
\vspace{-0.2cm}
\section{Empirical Experiments}
\label{sec:experiments} 
\vspace{-0.1cm}

\textbf{Benchmark Models}\footnote{Our models codes and data sources will be made public.} 

We benchmark our method with treatment effect estimation algorithms with the technique of estimating multiple outcomes with separate models. In this context, two mainstream methodologies in treatment effect estimation are \emph{meta-learners}~\cite{kunzel2017meta}~\cite{nie2017quasi} and causal trees and forests~\cite{wager2018estimation}. For each of these methods, we compare with the most representative algorithm known in literature, the quasi-oracle estimation algorithm, and causal forests. 

\emph{\textbf{Quasi-oracle estimation (R-Learner)}}. We use linear regression\footnote{\emph{SKLearn}'s ridge regression with zero regularization.} as the base estimator. Since the experiment treatments are randomly given, we use constant treatment percentage as propensity in the algorithm. We use the R-learner to model the gain value incrementality across treatment and control with an conditional average treatment effect function $\tau$ for each outcome dimension. Each sample in the test set is evaluated for the Individual Treatment Effect (ITE), and eventual metric is computed by combining ITE of all outcomes.  For instance, in the case of maximizing  Equation~\ref{eq:pstatement_objectives_1}, we would train an R-learner estimator for each of the $r$, $c$ and $q$ dimensions, then for each sample in the dataset, we compute the predictions for each of ITE $\tau_r$, $\tau_c$ and $\tau_q$, then compute score $s = 
\tau^{q}(\mathbf{x})(\tau^{r}(\mathbf{x}) - \lambda \tau^{c}(\mathbf{x}))$ for evaluation. 

\emph{\textbf{Causal Forest}}. We leverage the generalized random forest (\emph{grf}) library in R~\cite{wager2018estimation}~\cite{grflink}. For details, we apply causal forest with 50 trees, 0.2 as alpha, 3 as the minimum node size, and 0.5 as the sample fraction. We apply the ratio of two average treatment effect functions in ranking by training two causal forests. To evaluate matches with respect to the  effectiveness objective, we estimate the conditional treatment effect function for e.g. gain ($\tau_r$), cost ($\tau_c$), utility factor ($\tau_m$) i.e. train multiple Causal Forest models. For evaluation, similar as R-learner, we compute the score according to the pre-defined metric by combining ITE estimates in Equations~\ref{eq:pstatement_objectives_1}~\ref{eq:pstatement_objectives_2}. 
For hyper-parameters, we perform search on deciles for parameters \emph{num\_trees}, \emph{min.node.size}, and at 0.05 intervals for \emph{alpha}, \emph{sample.fraction} parameters. We also leverage the \emph{tune.parameters} option for the \emph{grf} package, eventually, we found best parameters through best performance on validation set\footnote{Best parameters we experimented: \emph{num\_trees}$=50$, \emph{alpha}$=0.2$, \emph{min.node.size}$=3$, \emph{sample.fraction}$=0.5$}. 

\emph{\textbf{Simplified CT Model}}. 
We use a simple parameterization to compute one score for any match as a benchmark model. To align with baseline and other methods in our experiments, we use a scoring function similar to logistic regression, i.e. $\sigma(\mathbf{w}^T [\mathbf{x}, \mathbf{y}] + b)$. Note $\mathbf{x}$ and $\mathbf{y}$ are feature sets for subject and matching candidates, respectively. The model is trained without weight regularization. We use the Adam optimizer with learning rate 0.01 and default beta values. We compute gradients for entire batch of data for optimization. For hyperparameter selection, variance in parameter initialization, we take the best validation set performance out of 6 random initializations. 

\emph{\textbf{Continuous Treatment Policy Matching Model (CTPM)}}. We implement our deep learning based models with Tensorflow~\cite{abadi2016tensorflow}. The graph construction utilizes Bayesian decomposition and implements subject model, bipartite embedding model, and continuous policy model, as well as normalizations using partition functions. We use two-layer neural networks with the same number of first-layer units in subject,  matching and policy models~\footnote{Number of hidden units is determined by validation results, 15 units for Ponpare dataset and 8 units for USCensus.}. Adam optimizer is applied with learning rate 0.01 and default beta values. The batch gradient is used to run the same number of iterations as simple CT model\footnote{Due to variance-bias trade-off across datasets, both CTPM and simple CT models are run for 2500 iterations for Ponpare dataset and 650 iterations for USCensus}. We take best validation set performance out of 6 random intitializations. 
\vspace{-0.1cm}

\textbf{Datasets} 

\emph{\textbf{Ponpare Data}} The Ponpare dataset is a public dataset with a coupons website~\cite{ponparekaggle}. The dataset is well-suited to evaluate our proposed methodology since it offers the discount levels of the coupons, which serve as continuous treatment intensities. Also the sessions in the dataset are when user browses the specific coupon, so offers treatment type to match with subject as the different coupons. The dataset contains session as row items where the instance contains customer, coupon browsed, a continuous discount percentage, whether or not purchase, and auxiliary features on customer and coupon. We leverage the open-source feature engineering code provided by~\cite{featureprepcoupon}.  The causal inference scenario focuses on estimating the combined benefits when we offer a continuous and variable discount percentage given a user-coupon match. We pre-filter the sessions where customers are below the age of 45. Due to disproportion of positive and negative samples, we subsample 4.0\% of sample of sessions that do not result in purchase. The eventual dataset is around $130,822$ samples, we utilize discount level as the continuous treatment policy, and use the median of the level to segment out sessions into treatment and control groups, indicated by binary variable $T$, Discount level is subsequently used as  continuous policy $\mathbb{P}$. For this dataset, we apply the optimization problem in~\ref{eq:pstatement_objectives_2} with $\tau_c$ as treatment effect for absolute discount amount with reference to cost, $\tau_r$ the purchase boolean variable with reference to benefit, and $\tau_m$ the geographical distance from user to the  product location for the coupon as extra cost related to delivery or travel. The $\lambda$ variable is chosen to be fixed at 0.1 across all models with the goal of adding distance factor into the objective. 

\emph{\textbf{US Census 1990}} The US Census (1990) Dataset (Asuncion \& Newman, 2007~\cite{uscensulink} contains data for people in the census. Each sample contains a number of personal features (native language, education...). The features are pre-screened for confounding variables, we left out dimensions such as other types of income, marital status, age and ancestry. This reduces features to d = 46 dimensions. Before constructing experiment data, we first filter with several constraints. We select people with one or more children (\emph{`iFertil'} $\geq$ 2)\footnote{\emph{`iFertil'} field is off-set by 1, \emph{`iFertil'}$=0$ indicating $\leq$ 15 year old male, \emph{`iFertil'}=1 no children.}, born in the U.S. (\emph{`iCitizen'} = 0) and less than 50 years old (\emph{`dAge'} $<$ 5), resulting in a dataset with $225,814$ samples. We select `treatment' label as whether the person works more hours than the median of everyone else, and select the income (\emph{`dIncome1'}) as the gain dimension of outcome for $\tau_r$, then the number of children (`iFertil') multiplied by $-1.0$ as the cost dimension for estimating $\tau_c$. The hypothetical meaning of this experiment is to measure the cost effectiveness, and evaluate who in the dataset is effective to work more hours. We apply optimization problem in   Equation~\ref{eq:pstatement_objectives_1} as comparison with Ponpare Dataset with $\tau_r$ as treatment effect in \emph{income}, $\tau_c$ as treatment effect in negative value of \emph{number of offspring}, and $\tau_q$ as effect on married or not as an overall weighting factor across the objective. This gives the objective hypothetical meaning of utility. The $\lambda$ variable is chosen to be fixed at 3.0 across all models to add a fixed cost weighting factor across income and offspring cost. 

For both datasets, we split training, validation and test with ratios 60\%, 20\%, 20\%. 
\vspace{-0.1cm}

\textbf{Evaluation Methodology} 

Our algorithm computes the matching score thus evaluate the \emph{differentiation} across instances. This means the average treatment effect of highly-scored instances will be better for the designed objective. We evaluate the algorithms in two ways. The first evaluation is Average Treatment Effect To Percentage (ATETP). This measure compute the effectiveness measure on the test data-set, then take an increasing percentage of the test set as to evaluate the average treatment effect according to the pre-defined causal metric in Equations~\ref{eq:pstatement_objectives_1}~\ref{eq:pstatement_objectives_2}. If the model scores the matches and treatment policies well, the ATETP should be high across the lower spectrum of percentages. We also use the ATETP area under curve (termed a-AUC) to be a numerical measure. The secondary metric is to plot a cost curve, i.e. to plot the treatment effect on reward $\tau_r$ versus cost $\tau_c$ as we increase percentage of coverage in the test set. This measure sees cost versus reward as the main concern, and we also compute the area under curve (termed c-AUC) to numerically measure performance. ~\footnote{For both a-AUC and c-AUC, the higher the measure the better.} 
\vspace{-0.1cm} 

\textbf{Experiment Results} 

\begin{figure}[h] 
  \centering 
  \includegraphics[width=0.80\linewidth]{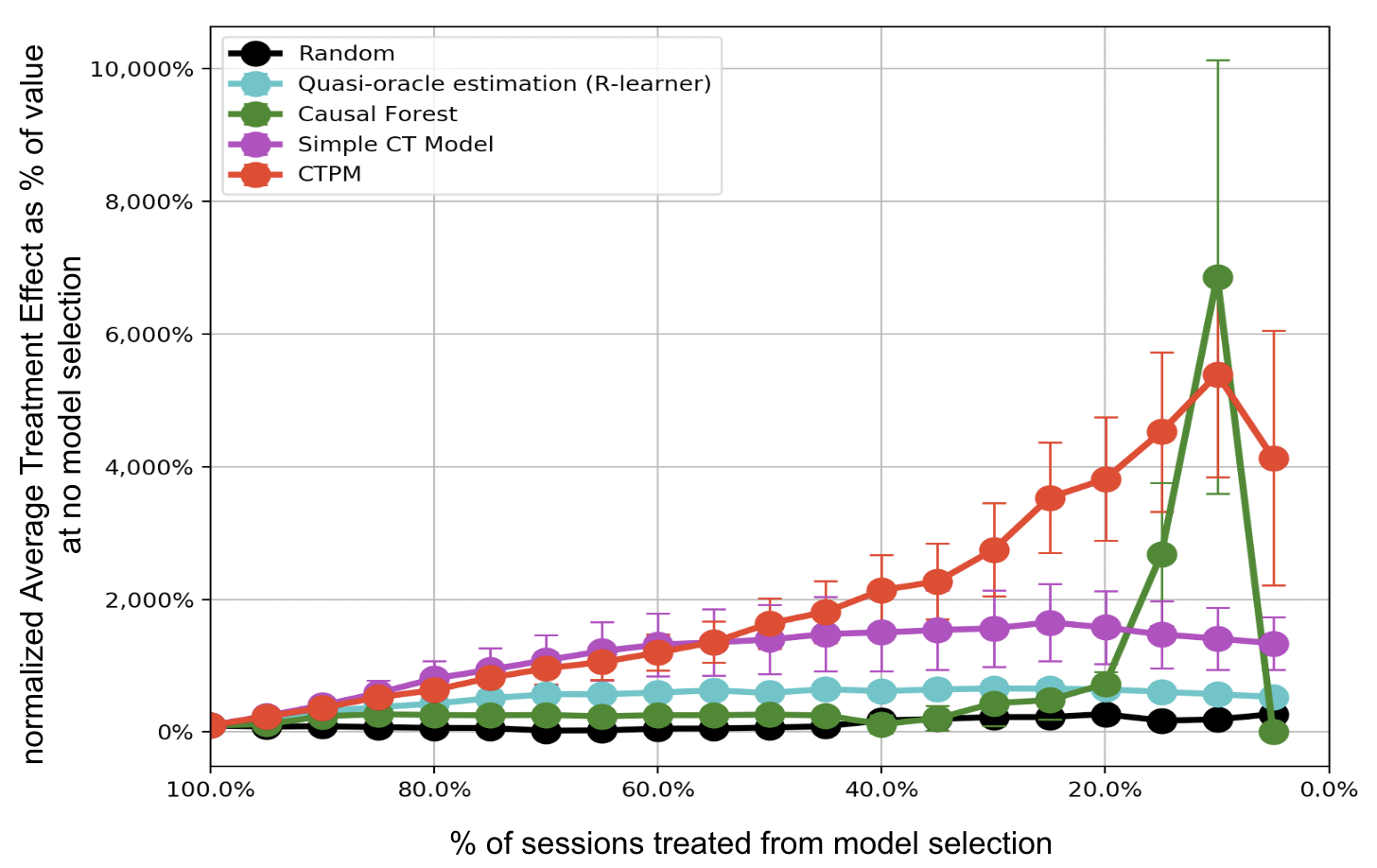} 
  \caption{Average treatment effect to percentage for Ponpare data.}
  \label{fig:Ponpare_test_ate} 
\end{figure}

\begin{figure}[h] 
  \centering 
  \includegraphics[width=0.80\linewidth]{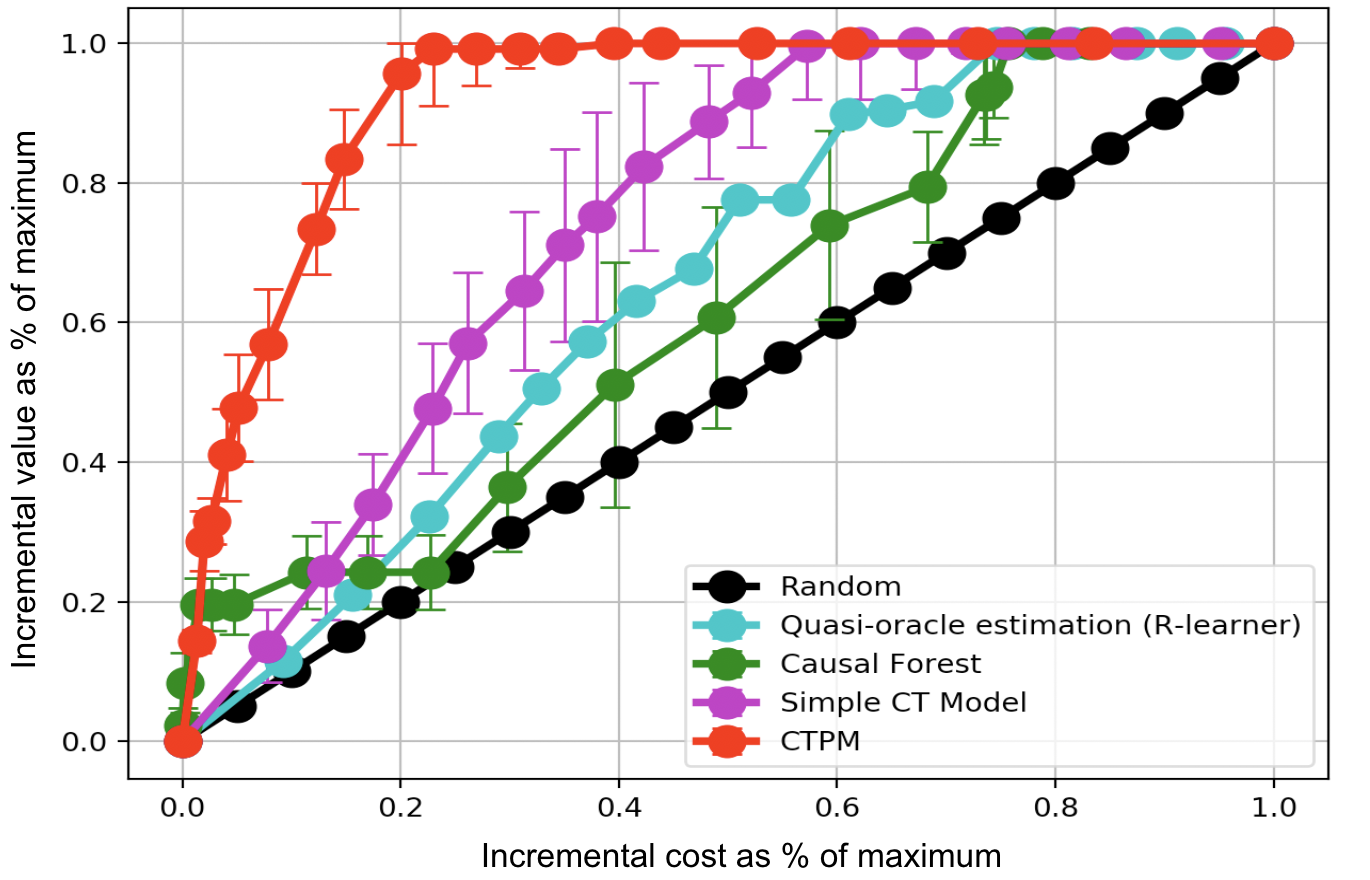} 
  \caption{Secondary measure cost curve for Ponpare data.} 
  \label{fig:Ponpare_cost_curve} 
\end{figure}

Figure~\ref{fig:Ponpare_test_ate} and Figure~\ref{fig:Ponpare_cost_curve} show results of causal models on Ponpare dataset\footnote{Standard deviations across 6 runs are indicated for both Ponpare and USCensus}. The simplified version of the model without continuous policy or bipartite matching is proposed by~\cite{du2019improve}. The CTPM out-performs R-learner, and simplied CT model on both ATETP curve and cost-curve. With peak at 10-20\% treatment, the CTPM produces ATE improvement at the most effective match instances across user and coupons. For cost curve, CTPM also outperforms other models. 
\vspace{-0.05cm}
\vspace{-0.05cm}

\begin{figure}[h] 
  \centering 
  \includegraphics[width=0.80\linewidth]{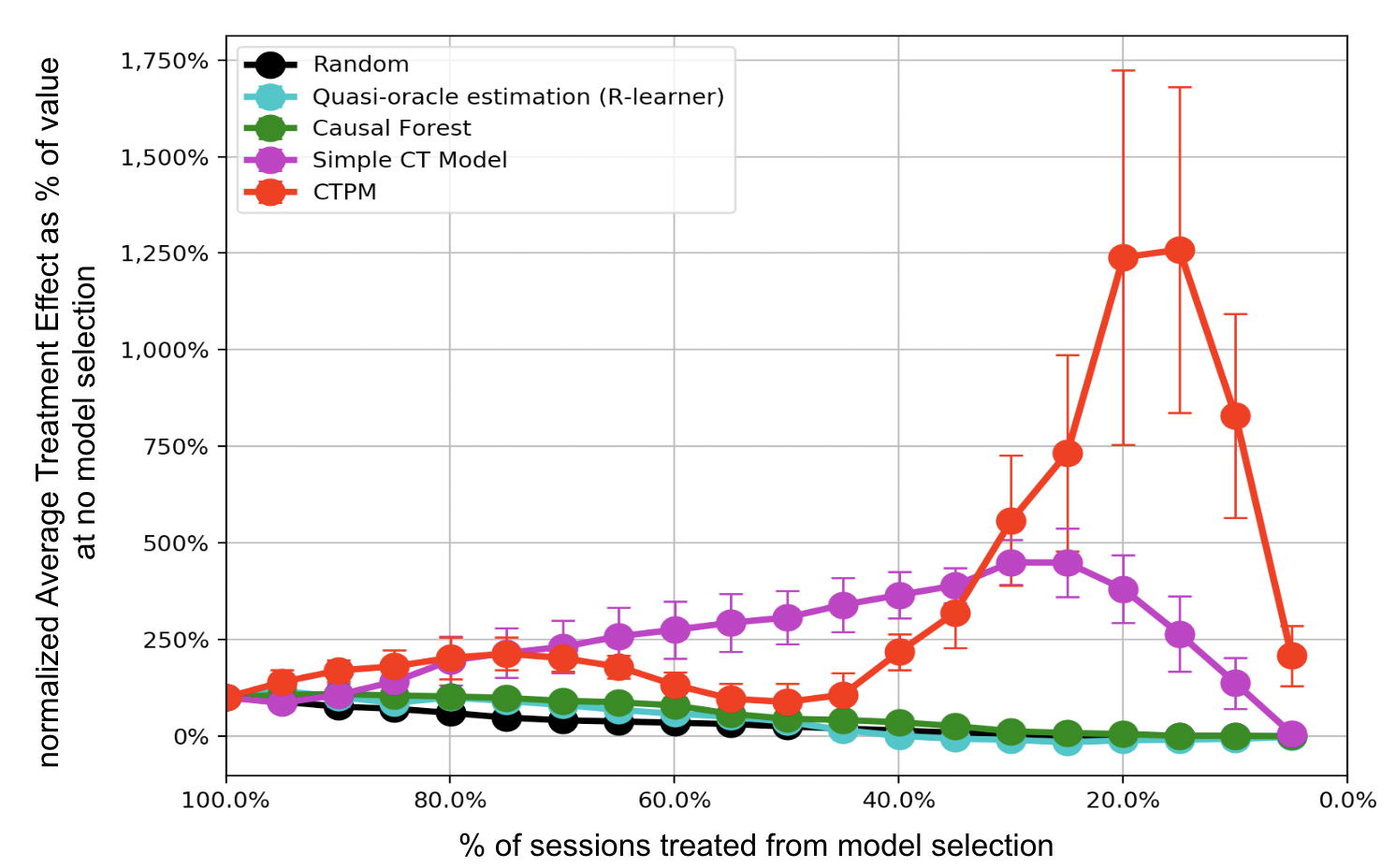} 
  \caption{Average treatment effect to percentage for US Census.} 
  \label{fig:uscensus_result_ate}
\end{figure}

\begin{figure}[h] 
  \centering 
  \includegraphics[width=0.80\linewidth]{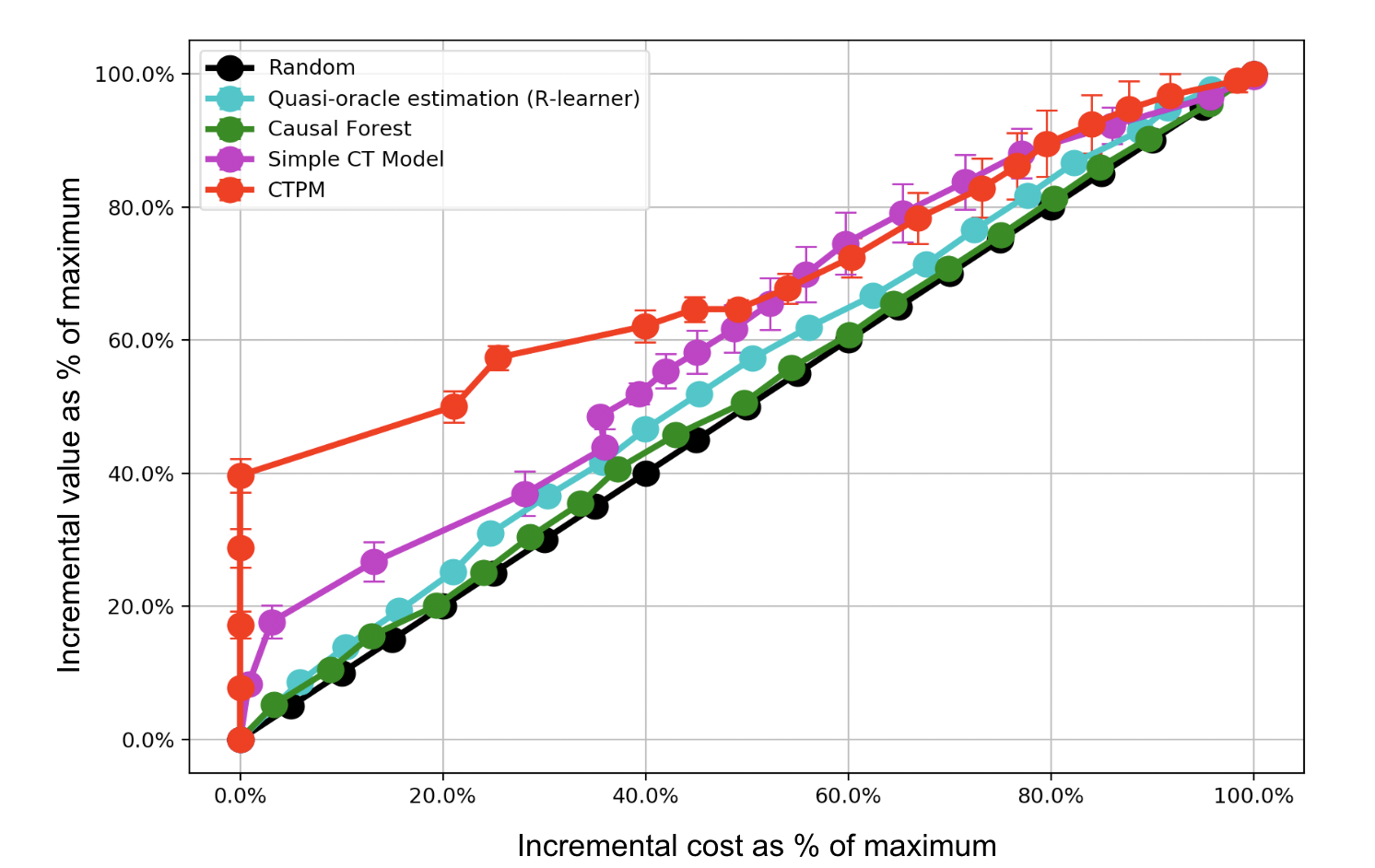} 
  \caption{Secondary measure cost curve for US Census.} 
  \label{fig:uscensus_cost_curve}
\end{figure}

Figure~\ref{fig:uscensus_result_ate} and Figure~\ref{fig:uscensus_cost_curve} show results of the CTPM on US Census. We observe higher ATE for the CTPM model in high-scored instances. CTPM could identify the most incremental instances without significant differences in cost. The model outperforms baseline R-learner and simplified CT model significantly both on the ATETP and Cost Curve measures. 
\vspace{-0.1cm}
\vspace{-0.1cm}

Table~\ref{tab:summary_result_table} summarizes results of the continuous treatment policy matching model. The CTPM outperforms prior models. On Ponpare dataset, CTPM out-performs R-learner by more than \emph{3$\times$} and improves 67\% upon Simple CT model in a-AUC. For c-AUC, CTPM improves 41\% upon R-learner and improves \emph{2$\times$} upon Simple CT model. On USCensus dataset, CTPM performs better than \emph{8$\times$} in terms of a-AUC than R-learner, and out-performs Simple CT model by around \emph{42\%}. CTPM is more cost effectiveness in terms of c-AUC by 28\% compared with R-learner, and 13\% improvement upon Simple CT model. 


\vspace{-0.1cm}

\begin{table}
  \caption{Summary of results across models and datasets.} 
  \label{tab:summary_result_table}
  \begin{tabular}{llllll}
    \toprule
    Algo/Dataset &  \multicolumn{2}{c}{Ponpare} & \multicolumn{2}{c}{USCensus}\\
    \midrule 
    Eval. Metric & a-AUC & c-AUC & a-AUC & c-AUC \\
    \midrule 
    Random &1.15 & 0.50 &0.31 & 0.50  \\
    \midrule 
    R-learner &5.06 & 0.65 & 0.40  &0.54\\
    Causal Forest &6.58 &0.61 & 0.53 &0.51 \\
    Simple CT &11.12 &0.74  & 2.47& 0.61 \\
    CTPM &\textbf{18.57} &\textbf{0.92}  & \textbf{3.51}  &\textbf{0.69}\\
    \bottomrule
\end{tabular}
\end{table} 

\textbf{Analysis and Interpretation} 

\begin{figure}[h] 
  \centering 
  \includegraphics[width=1.0\linewidth]{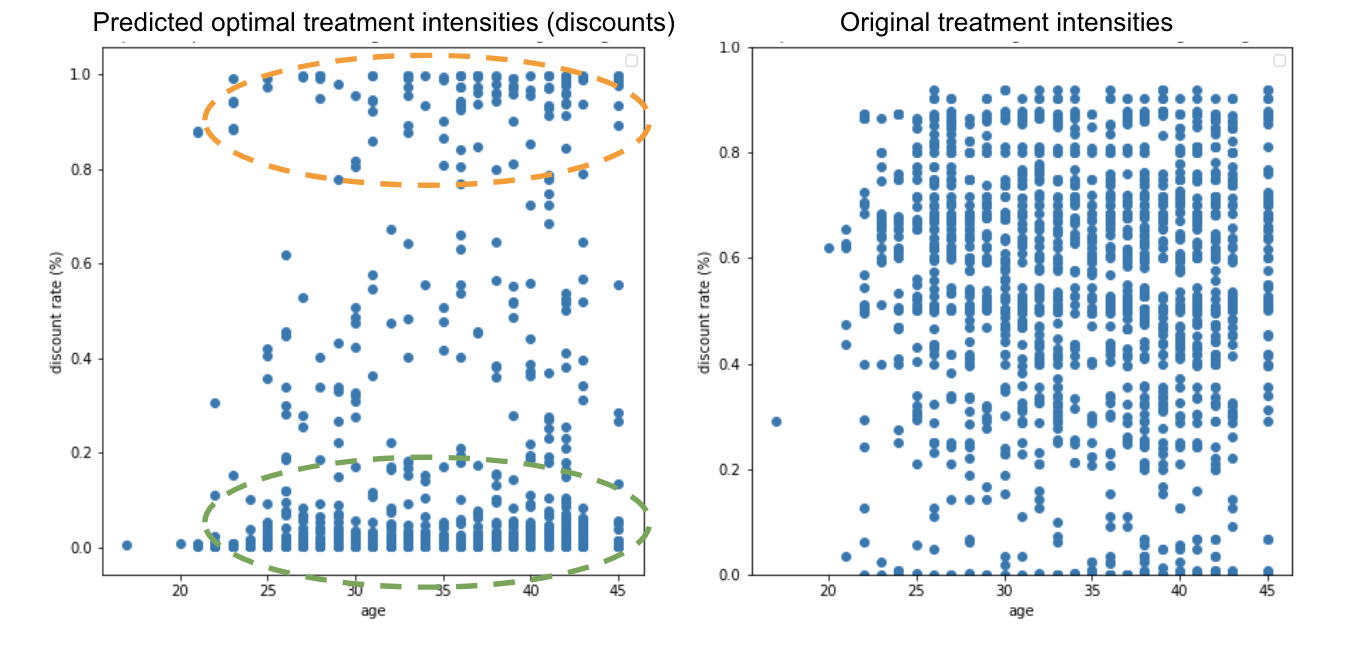} 
  \vspace{-0.2cm}
  \caption{Scatter plot comparison across optimal predictions from model (left) and original treatment intensities (right).}
  \vspace{-0.2cm}
  \label{fig:scatter_optimal_intensities} 
\end{figure}

The continuous policy model is able to predict the optimal treatment intensity. In Figure~\ref{fig:scatter_optimal_intensities}, we visualize the optimal discount per session for the genre \emph{`Health`} in the Ponpare test set. Compared with original treatment intensities, the optimal intensities from model prediction shows apparent segregation of low vs high intensity recommendations. This is shown by the data clusters near zero percentage (orange oval), and near full percentage (green oval). As age of the user increases, we see higher number of sessions with recommendation for high intensity of treatment. It implies older age group may have higher return for increased discount in the \emph{`Health'} genre, measured by the purchase probability. 
\vspace{-0.1cm}

\begin{figure}[h] 
  \centering 
  \includegraphics[width=1.0\linewidth]{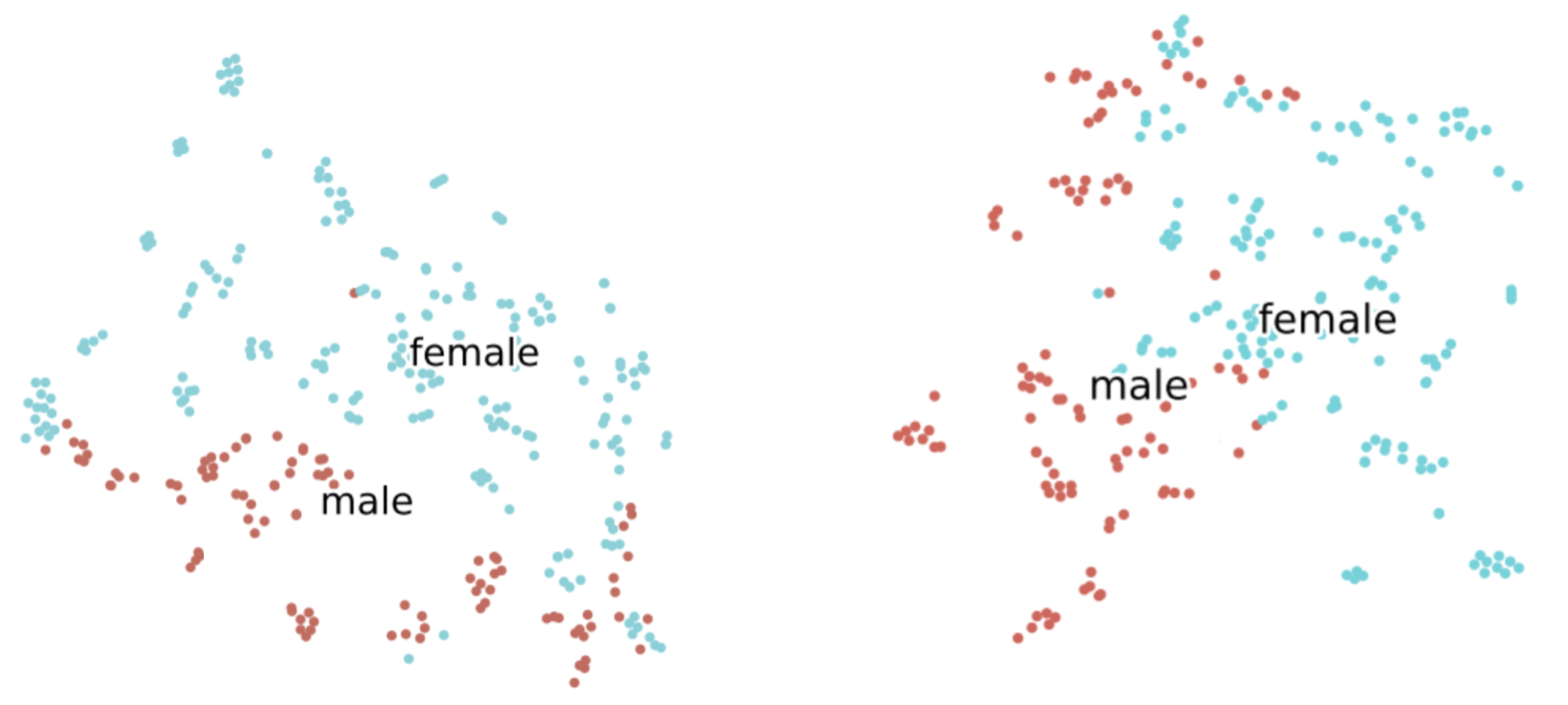} 
  \caption{Visualization of CTPM user embeddings for age group 30 (left) and 44 (right) from Ponpare dataset using T-SNE with color indicating user gender. } 
  \label{fig:user_emb_tsne2} 
\end{figure}

Figure~\ref{fig:user_emb_tsne2} shows the results of the learned embeddings by the causal models on the Ponpare dataset. The bipartite embedding space, in this case, is jointly learned across the treatment subjects (users who received coupons), and the treatment candidates (coupons). Figure~\ref{fig:user_emb_tsne2} plots the subject embeddings projected by the model, after running dimensionality reduction using 2D t-distributed stochastic neighbor embedding ~\cite{maaten2008visualizing} (t-SNE). We used a learning rate of 30 with a perplexity of 20. The idea is that the learned subject embedding normalized against different subject dimensions will be mapped to nearby points based on similarity in context and we can see that the learned subject embeddings are organized longitudinally by gender with two separable clusters. 

\vspace{-0.1cm}

\section{Conclusion and Discussion} 
In this paper, we proposed a model that combines methodologies to utilize continuous space treatment policy, and matching treatment subjects to candidates. These methods categorize the intensity and complexity of treatment action space thus significantly improve the performance of decision models. The models are able to make better decisions that maximizes holistic average treatment effects and on cost versus reward effectiveness. Further, the CTPM is able to predict the optimal treatment policy per matching instance, based on the contextual features. Finally, the matching algorithm offers latent variable models with embedding space to characterize the joint space in the subject and candidate instances and further improve results on applicable datasets. 

The proposed algorithms holistically optimize with respect to action spaces, for a flexible objective combined with multiple treatment effect functions. For future work, our proposal offers potential to combine with other deep learning techniques such as sequential, recurrent models, generative models, and can be potentially extended and applied to other scientific domains.





\bibliography{main}
\bibliographystyle{icml2019}

\end{document}